\documentclass[letterpaper]{article} 

\ifcsname arxiv\endcsname
\usepackage{Style/arxiv}
\else
\usepackage[submission]{Style/aaai25}
\fi
\usepackage{times}  
\usepackage{helvet}  
\usepackage{courier}  
\usepackage[hyphens]{url}  
\usepackage{graphicx} 
\usepackage{natbib}
\usepackage{caption} 
\usepackage{algorithm}
\usepackage{algorithmic}

\usepackage{newfloat}
\usepackage{listings}
\DeclareCaptionStyle{ruled}{labelfont=normalfont,labelsep=colon,strut=off} 
\floatstyle{ruled}
\newfloat{listing}{tb}{lst}{}
\floatname{listing}{Listing}
\ifcsname arxiv\endcsname
\usepackage{xcolor}
\usepackage{colortbl}
\usepackage{comment}
\usepackage[hidelinks]{hyperref}
\fi
\usepackage{booktabs}
\usepackage{amsmath}
\usepackage{amssymb}
\usepackage{bm}
\usepackage{subcaption}
\newtheorem{definition}{Definition}
\newtheorem{claim}{Claim}
\newtheorem{observation}{Observation}

\title{A theory of understanding for artificial intelligence: composability, catalysts, and learning}

\author {
    Zijian Zhang\textsuperscript{\rm 1,\rm 2},
    Sara Aronowitz\textsuperscript{\rm 1},
    Al\'an Aspuru-Guzik\textsuperscript{\rm 1,\rm 2,\rm 3,\rm 4,*}
}
\affiliations {
    \textsuperscript{\rm 1} University of Toronto
    \textsuperscript{\rm 2} Vector Institute
    \textsuperscript{\rm 3} Lebovic Fellow
    \textsuperscript{\rm 4} Canadian Institue for Advanced Research\\
    \textsuperscript{\rm *}\texttt{alan@aspuru.com}
}

\begin{document}
\maketitle

\begin{abstract}
Understanding is a crucial yet elusive concept in artificial intelligence (AI). 
This work proposes a framework for analyzing understanding based on the notion of composability. 
Given any subject (e.g., a person or an AI), we suggest characterizing its understanding of an object in terms of its ability to process (compose) relevant inputs into satisfactory outputs from the perspective of a verifier. This highly universal framework can readily apply to non-human subjects, such as AIs, non-human animals, and institutions.
Further, we propose methods for analyzing the inputs that enhance output quality in compositions, which we call catalysts. 
We show how the structure of a subject can be revealed by analyzing its components that act as catalysts and argue that a subject's learning ability can be regarded as its ability to compose inputs into its inner catalysts.
Finally we examine the importance of learning ability for AIs to attain general intelligence. Our analysis indicates that models capable of generating outputs that can function as their own catalysts, such as language models, establish a foundation for potentially overcoming existing limitations in AI understanding.
\end{abstract}

\section{Introduction}

Understanding is both a technical term in philosophy and a topic of interest in science and everyday life. In philosophy, discussions of understanding often center around its relationship with other epistemological concepts, as well as its connection to epistemic and practical norms~\cite{baumberger2016understanding, hannon2021recent}. From this perspective, a successful definition of understanding matches pre-theoretic intuitions, draws links with concepts like knowledge and belief, and explains why understanding is desirable. With a satisfying definition in hand, we can then ask whether machines understand. 

However, in the pursuit of machine intelligence, the question of understanding has arisen in a different, and urgent form: in a passive sense, we want to know whether current AI systems can understand certain objects, such as morality, psychology, or Newtonian physics. For example, \citet{krenn2022scientific} proposed potential tests to verify whether a system has scientific understanding and highlighted the importance of whether AI can transfer their understanding to humans. In an active sense, creating AI systems that can understand is also an important aim of many designers. Among various efforts in building AI systems, the advent of large language models (LLMs) \cite{brown2020language, openai2023gpt4, team2023gemini, anthropic2024claude} has dramatically advanced this field by providing a universal language processor that is powerful yet not limited to specific tasks. Many of these models \cite{chowdhery2023palm, sun2021ernie} have been reported to outperform humans in predefined natural language understanding challenges \cite{wang2018glue, wang2019superglue} while maintaining generalizability to many other tasks. This generalizability has enabled the construction of LLM-based agents that show unprecedented universal ability in observing, planning, and acting \cite{wang2023survey}.

Despite the advancement, controversy still surrounds whether LLMs truly understand \cite{mitchell2023debate, browning_ai_2022, firt2023artificial, lenci2023understanding}, which is closely related to some of the practical problems that LLMs currently face. One of the well-known issues is hallucination, where LLMs generate nonsensical information \cite{ji2023survey, shen2023chatgpt}. Although humans also generate incorrect information, LLMs additionally suffer from the hardness of model editing \cite{yao2023editing, hoelscher2023detecting, zhang2024comprehensive}, making it difficult to remove hallucinations by updating knowledge.
In addition, LLMs constantly have much lower performance on counterfactual tasks that rarely appear in the training set \cite{wu2023reasoning}, implying that LLMs are far from the kind of ideal thinker who can generate and use general-purpose concepts for out-of-distribution problem solving. 
LLMs also face the problem of prompt brittleness, in which even trivial changes in input (prompt) can drastically alter the output produced by LLMs \cite{lu-etal-2022-fantastically, min2022rethinking}. 
Further, a more general problem lies in challenges for statistical learning as a path to understanding: the training regime of language models may not be able to ground understanding, even for an arbitrarily sophisticated machine \cite{bender2020climbing}. 

All these problems imply that LLMs are still far from achieving understanding at the human level. This understanding would enable them to operate their memory robustly, link observations to concepts, and produce unusual yet correct outputs. However, it has been argued that LLMs have some significant ability to understand and produce meaningful utterances according to some standards \cite{mandelkern2023language}. Further, many higher-level constructions, such as retrieval augmented generation (RAG) \cite{lewis2020retrieval, gao2023retrieval} and LLM-based agents \cite{wu2023autogen,li2023camel,wang2023survey, xi2023rise}, have been built on LLMs and may be closer to understanding. Analyzing the understanding achieved by these composite systems presents another layer of complexity. More generally, pessimism about LLM understanding can raise the bar for understanding so high that some humans and animals may not meet it. 

Here,  we introduce a framework to analyze understanding with the notion of composition, which involves synthesizing inputs into an output in the most general sense. 
We propose to characterize the understanding of an object by the subject's ability to properly compose inputs that a verifier believes to be appropriate. This framework allows for accessing understanding according to a broad range of behaviours beyond traditional approaches dependent on concepts such as beliefs, concepts, and assertions. 
With this method of characterizing understanding, we analyze the understanding of current AI systems from the aspects of universality and scale. 

In order to go beyond a black-box behaviourism, we propose using the concept of ``catalyst'' to capture assistance in understanding, whether physically internal or external to the subject. The ability to produce catalysts can be used to characterize the subject's learning ability.
Finally, we use the view of learning ability to view the problems of LLMs we mentioned above. We will provide arguments that imply that LLMs are still far from general intelligence, and yet how future
LLM-based systems may offer promising solutions to fill the gaps.

\section{What is understanding?}
\label{sec:whatisunder}
We start by noticing the challenges in developing a general account of understanding. Consider the following uses of ``understanding'':
\begin{itemize}
\item Alice understands Newtonian mechanics. (a theory)
\item An LLM understands black holes. (an object)
\item Charlie understands that the sky is blue. (a proposition)
\end{itemize}
These examples show that, at least intuitively, understanding can apply to objects and subjects of many types. This implies that a general theory of understanding must apply universally to these objects and subjects. However, achieving this desired universality introduces two complexities arising from both the diversity of objects and subjects. 

To illustrate the complexity introduced by the diversity of objects, let us explore the concept of ``knowing'' in relation to ``understanding''. While ``knowing'' can be predicated of any true proposition, ``understanding'' carries a more complex implication about its object. To see this, consider the following examples:
\begin{itemize}
    \item ChatGPT knows that the sky is blue.
    \item ChatGPT understands that the sky is blue.
    \item Bob knows his phone number.
    \item Bob understands his phone number.
\end{itemize}
When the verb ``knows'' is replaced with ``understands'', the sentences seem to imply a substantially different meaning rather than merely increased confidence or strength. When we say ChatGPT understands that the sky is blue, without further context, it implies that ChatGPT possesses more background knowledge about this fact. For example, it knows that the color of the sky is caused by the scattering of photons by particles in the atmosphere. Similarly, when Bob understands his phone number, it sounds odd unless we can see in context that he does not merely know his phone number but also grasps some additional features of it, such as how to help others remember it easily or how to dial it from abroad.
Understanding seems to always imply some additional ability related to the understood object. While it may be easy to infer what ``understanding'' implies for a specific object, it is challenging to define a universal rule that specifies which additional abilities should contribute to the understanding of an object.

Most contemporary philosophers sidestep the range of objects of understanding by focusing on propositional understanding, where the object of understanding is limited to propositions (for instance \citet*{Khalifa2012-KHAIUO} or \citet*{Sosa2010-SOSKFW-2}, see \citet*{hannon2021recent} for a review). 
While this approach might be suitable for debates about the connection between understanding and other epistemological concepts, such as knowledge, it is insufficient in our context: in the case of AI, we have a direct interest in whether they can understand certain domains or subject areas and are only indirectly concerned with whether they understand particular propositions. For instance, \citet*{bender2020climbing} survey discourse about whether LLMs can understand ``how human beings communicate'', ``common sense'', ``running prose'', and so on (p5186).  

Another complexity of understanding is rooted in the diversity of its subjects. Consider the following example:

\begin{quote}

\textbf{(Understanding with assistance)}  Bob is not good at explaining or identifying safe driving. However, when he drives a car equipped with a safety system, he adheres to the system's prompts and has an excellent driving record. Though it might contain many rules for safe driving, the system itself cannot drive without Bob. 
\end{quote}

Does Bob understand safe driving? He surely does not understand it in the same way that someone who can drive safely without the system does. 
One way to deal with this case would be treating Bob and the safety system as a composite subject. 
But this approach ignores the useful structure inside the composite subject: Bob's way of understanding is in relation to a particular aid, the safety system. In fact, we find it an underappreciated feature that understanding is very often significantly dependent on an aid: understanding a multiplication table from memory, by using an algorithm, or by relying on notes are importantly different. This second complexity is in the subject of understanding.

What kind of account of understanding considers the broadness and complexity of the objects and subjects? We contend that the answer is one that is both \textbf{practical} and \textbf{minimal}. A \textbf{practical} account of understanding links understanding to behaviours of all kinds: when a subject understands a friend in this practical way, it means they can act in ways that cheer up the friend when they are feeling sad and can also remember the friend when encountering news that particularly interests them. 
That is, rather than practical and theoretical understanding being disjoint categories, we here use practical understanding to encompass all kinds of responses, including, but not privileging, belief and other mental states. Practical accounts can explain why understanding attaches to all kinds of objects (i.e. because behaviours can concern places, things, and so on, in addition to propositions), as well as the implication that understanding involves not just grasping a single fact (i.e. because saying ``understands'' when we could have said ``knows'' implies further cognitive or behavioural consequences).

A \textbf{minimal} account of understanding means that it imposes few constraints on the subject's inner structure. Imagine that Alice understands physics with a mnemonic in her brain and David understands physics with a mnemonic written in a notebook. We contend that combinations including Alice with her memory, Alice without her memory, David with his notebook, and David without his notebook can all be considered as subjects of understanding depending on the context. 
On a non-minimal theory, determining a subject is a substantial matter that admits a restrictive class of things. Further, many accounts of understanding assume that the subject must be a person, which is a category that is presumably defined by external, metaphysical considerations. In contrast, a minimal account treats the subject definition as a flexible part of the analysis of understanding. In Section~\ref{sec:compo}, we make this idea more precise.

Does a minimal, practical theory treat understanding as a black box? In other words, would our account of understanding focus solely on the subject's behaviour, ignoring their internal structure? That is, if we treat Bob and the safety system as a composite subject, would we ignore the difference between the composite system and a person who behave the same without assistance?
Surprisingly, the answer is no: using the strategy of subject decomposition in Section~\ref{sec:structure}, we will aim to get into the subject's structure while still defining understanding in a way that can be verified from the outside.

\section{Understanding as composability}
\label{sec:compo}
\begin{figure*}
    \centering
    \includegraphics[width=1\linewidth]{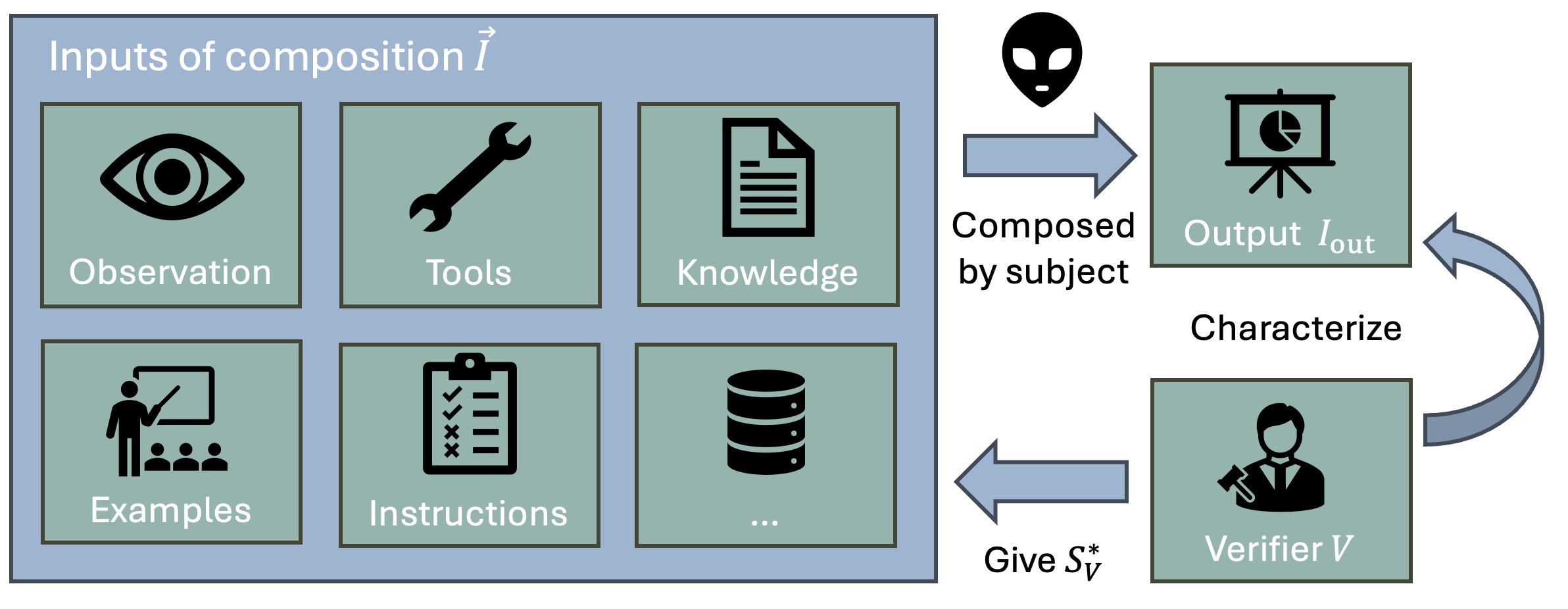}
    \caption{A diagrammatic depiction of composition and the verifier of understanding. We define composition as a general process that produces an output based on inputs, where the inputs can be any reasonable entity in the verifier's view. In our framework of understanding, verifiers characterize a subject's understanding of an object $O$ by its outputs in composing inputs that are related to $O$. }
    \label{fig:comp}
\end{figure*}

To design our practical, minimal theory of understanding, we need a theoretical tool to describe the subject's behaviour. These two requirements suggest the following definition:

\begin{definition}[Composition]
Composition is a process in which a subject creates the output $I_{\rm out}$ given a list of inputs $\vec{I}=[I_0, I_1, ...]$.
\end{definition}

In the definition, the ``inputs'' and ``output'' of a composition can be \textbf{anything} the theory user believes to be reasonable in the context. For example, consider syllogistic reasoning. Imagine a person facing statements: ``Deduce using syllogism'', ``All men are mortal'' and ``Socrates is a man''. A natural answer from the person would be, ``Socrates is mortal''. From the notion of composition, we regard the person as composing the first two statements into the conclusion. 
The inputs of compositions are not limited to a certain category. For example, the inputs of compositions can also include tools and instructions, with the outputs being the results of using these tools. Consider a search engine and an instruction to write a report about apple. The process of using the search engine to gather information and produce the report can be viewed as a composition, where the inputs are composed into the finished report. This broad definition of inputs and outputs makes the concept of composition flexible enough to describe broad behaviours related to understanding, which helps make our theory of understanding \textbf{practical}.

Further, composition is also a general concept that can apply universally to subjects. For example, Turing machines can be treated as subjects that compose inputs from the tape, and the processed state of the tape can be regarded as the output of composition \cite{turing2009computing}. More abstractly, we can treat the universe as a huge machine that composes its current state, including the information of all the elementary particles, into its next state. This universal applicability of composition helps to make the theory subject-general and therefore \textbf{minimal}. 

We continue by discussing how to use composition to build an operationalization of understanding. Here, we introduce the concept of a \textbf{verifier} of understanding. A verifier of understanding is a subject that can assess instances of understanding (which might be their own understanding), including comparing understanding across subjects or contexts, and defining which inputs to composition are related to which objects of understanding. A reader who prefers a more objective theory of understanding that is not relativized to a verifier could consider the verifier as a placeholder for the role of a further epistemological theory -- in this case, we are defending the skeleton of an account of understanding which requires further specification to be a final, objective theory. With this construction, we formally give the core definition introduced by this work:

\begin{definition}[Understanding as composability]
\label{under-as-comp}
A subject's understanding of an object $O$, in the view of a verifier $V$, can be fully characterized by the set $S$, which contains all the tuples $(\vec{I}, I_{\rm out})$, where $\vec{I}$ is from the set $S_V^*$ of all the lists of inputs that $V$ deems to be related to $O$ and $I_{\rm out}$ is the output of the subject composing $\vec{I}$ .
\end{definition}

We use an example to illustrate how the definition can be used for ascribing understanding in context.
\begin{quote}
\textbf{(Understanding phenomena)} A verifier can evaluate Charlie's understanding of the phenomenon ``the sky of the earth is blue'' by his output $I_{\rm out}$ of composing inputs that the verifier deems related (i.e., in $S^*_V$), such as ``what causes the sky to be blue?''. Some sophisticated verifiers may include ``I want to see the blue sky of the moon'' in $S^*_V$ because it tests whether Charlie can realize this is impossible.
\end{quote}
Our definition does not give criteria for understanding. Instead, the definition aims to provide a framework for analyzing instances of understanding by the set $S$ that includes all behaviours of the subject about the understood object $O$. Our framework is minimal as it does not require any extra property of the subject besides doing compositions. As in the Turing test, our definition emphasizes the role of the verifier in making a judgment, in which the verifier is also a subject who may have no more abilities or theoretical privilege than the subject whose understanding is verified. In the above example, we can see how the verifier's assessment of understanding can vary with their own understanding of the object. Therefore, in some cases, the verifier's judgement fails due to limitations of their own knowledge.
\begin{quote}
\textbf{(Mistaken understanding)} An English speaker may believe that Claude \cite{anthropic2024claude}, an LLM, understands English when it composes English sentences as inputs and produces reasonable reactions as outputs. However, a person who cannot tell English from French may believe that Claude understands English even when the person only sees Claude processing French. This failure is rooted in the person's flawed internal set of $S^*_V$ and then the judgment of the composition outputs. 
\end{quote}

Misunderstanding understanding is not rare in daily life, and similar situations may become even more important in the AI era when people might not be able to verify the answers stemming from AI tools \cite{ji2023ai, superalignment}. 
Different verifiers, with different types of understanding, may make totally different judgments about a subject's understanding of an object. Our framework, being minimal, allows the analysis of the verifiers using the same method as the subject. For example, to test a verifier's understanding of being a verifier with respect to a class of objects, we might ask it whether a certain list of inputs is related to those objects and why. Answering such questions can be regarded as a composition based on the verifier's understanding. 

Finally, let us use LLMs as an example and analyze the extent and limitations of their understanding of apples. First, in the view of a human verifier, inputs related to the understanding of apples might include questions such as ``What is an apple?'' and ``What can you cook with apples?''. LLMs, which are supposed to process text inputs, usually respond properly to these questions. However, inputs, such as a cup of apple juice and the sound of eating an apple, are also related to the understanding of apples from the perspective of most human verifiers. Humans are expected to drink apple juice and tell that it is made from apples, which is unachievable by LLMs. In this way, we find a limitation in LLMs' understanding. Our framework, in addition, emphasizes that the verifier of understanding also has limitations. It is usual that humans wrongly believe LLMs' claims because of their confident tone. It is possible that LLMs present incorrect information about apples and fool humans by mimicking agricultural experts.
 
\section{Characterizing understanding}

If you have read the book \textit{Dr. Jekyll and Mr. Hyde}, could you draw the layout of Dr. Jekyll's house? The writer Vladimir Nabokov used to have his students produce drawings like this as part of a literature course (see Figure~\ref{fig:house}). 
These drawings are tests of understanding: Nabokov used both to test his students and to make them aware of the limits of their own understanding \cite{nabokov2017lectures}. You could also ask an artificial agent to draw the house: would it succeed, and if so, what would that tell us?

In Section~\ref{under-as-comp}, we have shown how to use the set $S$, which describes all the related behaviours of the subject, to characterize its understanding of an object. Commonly, the preference of understanding varies from verifier to verifier. It is hard to give a verifier-independent method to characterize instances of understanding. However, we can show that there are still useful heuristics for analyzing a set $S$ and the corresponding understanding. In the following, as examples, we demonstrate how the notion of composability can help us find two useful and general characteristics of instances of understanding.

\subsection{Universality}

\begin{figure}
    \centering
    \includegraphics[width=0.45\textwidth]{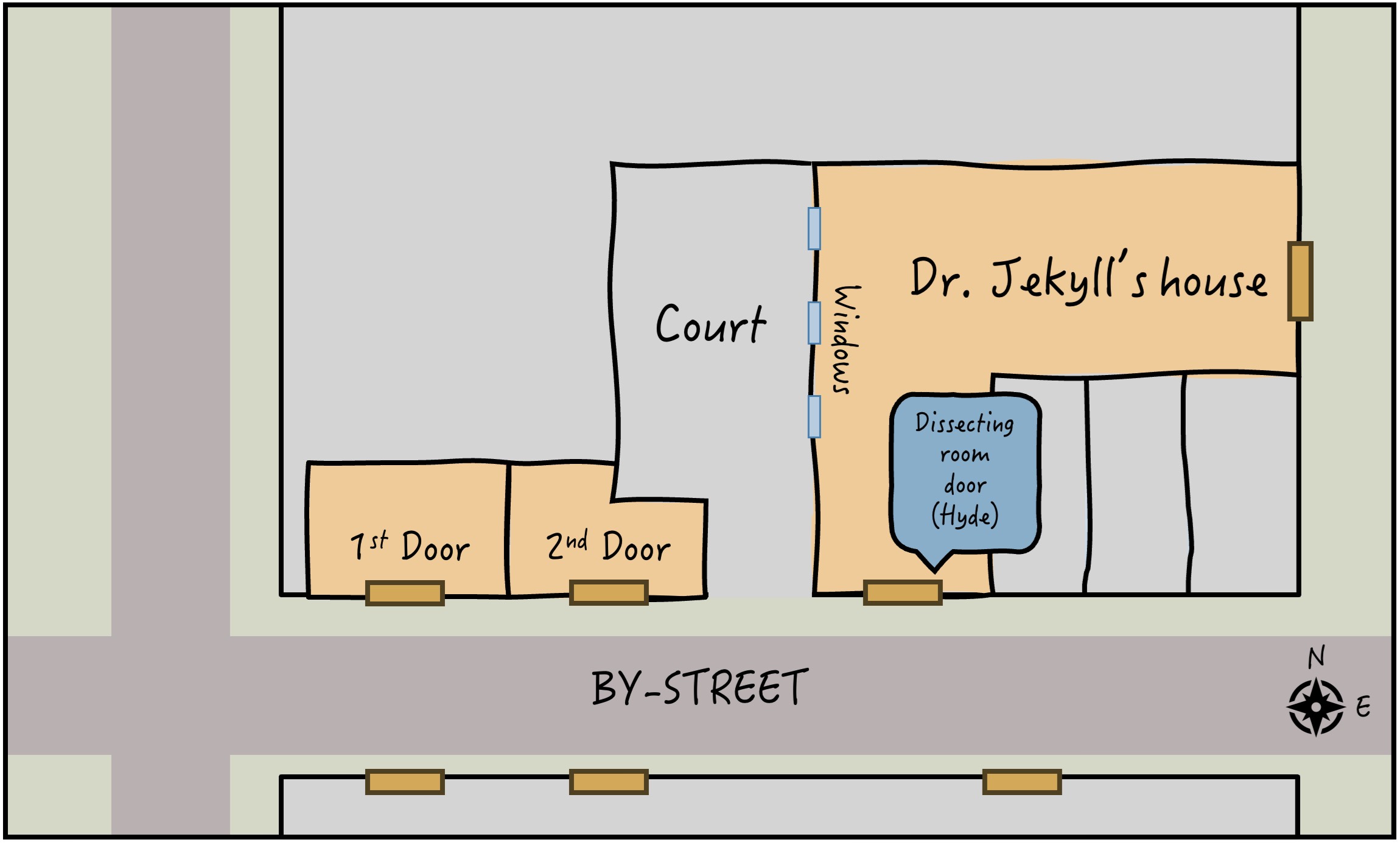}
    \caption{A replication of a student's drawing of the layout of Dr. Jekyll's house with Nabokov's annotations. \cite{nabokov2017lectures} The student demonstrates an understanding of reading by composing the novel into a diagram. The diagram, as well as the process of making it, also helps the student understand the novel.}
    \label{fig:house}
\end{figure}

Humans can input and output many different types of information. One dimension of variation is modality: verbal, auditory, visual, tactile, etc. In the context of verbal inputs, humans can also process multiple types of natural or programming languages. 
This ability allows humans to compose multi-type inputs that are related to the object when they understand it, as well as to produce multi-type outputs. 
For example, humans can read a novel and draw the scenarios in it, as in the example shown in Figure~\ref{fig:house}. It is also possible to use the drawing as an input to help human readers understand the novel better.
As we have demonstrated how a verifier's understanding can influence their assessment of understanding, it is easy to see how the generality of human processing is a significant factor in our perception of a particular instance of understanding. This generality is not fully universal, but we can infer that approaching universality would similarly increase and alter understanding.

It is straightforward to observe that current AI systems, as of 2024, have not reached the same level of universality as humans. This implies that these AIs' understanding (1) is different from humans, though they might outperform humans in some way; (2) falls short in composing many inputs that human verifiers might find critical. Machine learning models are not new to outperform humans in their tasks, such as playing games \cite{silver2016mastering, vinyals2019grandmaster}, image recognition \cite{he2015delving}, and some natural language processing tasks \cite{chowdhery2023palm, sun2021ernie}. Although these models acquired an $S$ that humans cannot achieve, they are still accused of not understanding the tasks \cite{Michael2016AlphaGo, Tanya2016AlphaGo, mitchell2023debate}.
To see why these criticisms are reasonable, consider AlphaGo and related models \cite{silver2016mastering, silver2017mastering}, which are intrinsically unable to process natural languages, whether as inputs or outputs. This makes them fundamentally unable to explain why a particular move is better than another in a way that makes the understanding transferable to humans, though they might possibly produce some numerical evidence showing their move is reasonable.
We note that this analysis also applies to most computer vision models and even models for specific natural language processing tasks.

The above observations make it easier to see why the increased universality of LLMs can be seen as an important advancement. LLMs are designed to accept input texts and even images with different types and purposes. It is possible to specify instructions, examples, and background knowledge to help LLMs finish complicated tasks and produce human-friendly outputs. From the view of human verifiers, this makes LLMs able to compose information similarly to humans and thus produce more satisfactory outputs to the elements in $S^*_V$. 
Therefore, we see that LLMs, though probably not strictly outperforming some specific models, offer a way to produce agents with more favourable instances of understanding for human verifiers. We also note that the importance of the transferability of AI understanding, as discussed in \cite{krenn2022scientific}, is also driven by the preferences of human verifiers.

Besides being human-friendly, LLMs' increased universality probably brings a more critical advantage: their ability to use tools. Tool use has been already regarded as an important demonstration of human intelligence \cite{vaesen2012cognitive}. LLMs, as being universal in processing languages, are shown to be able to use various tools \cite{wang2023survey, xi2023rise}, such as RESTful APIs \cite{qin2023toolllm}, domain expert systems \cite{ge2024openagi}, and other machine learning models \cite{yang2023mm}. It is not hard to imagine that an LLM-based agent can prepare inputs in a structured language (e.g., JSON) to machine learning models (e.g., AlphaGO) and interpret the outputs to humans in English text. We can also equip LLMs with a text-to-speech model that accepts JSON inputs to make their outputs more human-like. In contrast, models that handle a specific task generally cannot use these tools, though they might perform better than more universal models. This makes it difficult for them to improve their understanding without undergoing laborious model training.

\subsection{Scale}

Another aspect we consider central to the subject's understanding of a subject is the scale of inputs and outputs it can handle.
For example, suppose Charlie could only compose information from articles from his English textbook up to 200 words, and, after one week of study, Charlie could compose with inputs of much longer English articles. In this case, it is safe to say Charlie understands English better after one week. 
Similar cases appear when it is expected that subjects who can handle large-scale bodies of information can also handle small ones with similar complexity. 
For example, to evaluate an LLM's understanding of reading comprehension, it is reasonable to test whether it can produce satisfactory outputs on longer texts.
Though larger bodies of information are not always harder to compose than small ones, it is usually useful to check the scale of the inputs in $S$.

Including scale as a special dimension seems redundant because one may regard large-scale input as just another type of inputs. We admit this ambiguity but suggest that the scale is of special importance due to its inherent difficulty.
We point out that handling large-scale information is especially challenging for both humans and AI. 
Though being universal in input types, most LLMs are constrained by a limited input size.
GPT-3.5 is capable of handling only up to 4,096 tokens in the initial version (each word takes one or more tokens). 
Upon its release, the more advanced GPT-4 is restricted to a mere 8,192 tokens \cite{openai2023gpt4}. 
This limitation is due, in part, to inherent structural limitations within the transformer neural network. 
Some LLMs claim to be proficient with longer inputs, but empirical studies indicate a consistent decline in performance as input lengths increase \cite{liu2023lost, shi2023large, li2024long, li2023long}. 
In contrast, humans can process long inputs, including long articles or videos, though they may process them much more slowly than LLMs.
From this perspective, humans demonstrate a greater understanding in processing and handling articles and watching videos. 

\section{Structure of understanding}
\label{sec:structure}

In the preceding sections, we proposed a framework of understanding that is minimal and independent of the properties of the subject. This independence makes it possible to define subjects with flexibility when describing a situation involving understanding.  This allows us to handle cases, such as the case of assisted safe driving of Bob in Section~\ref{sec:whatisunder}, where the subject of understanding could be composite. Our framework, focusing on the operational implication of understanding, has no difficulty in admitting the composite system as the subject of understanding. Further, by emphasizing ``composing'', our framework also provides other interpretations of Bob's safe driving:
\begin{itemize}
    \item Bob makes proper outputs by composing the safety system, the car, and other inputs. 
    \item The safety system makes proper outputs by composing Bob, the car, and other inputs.
    \item The car makes proper outputs by composing Bob, the safety system, and other inputs.
\end{itemize}
We allow all three interpretations and hold that each provides insights into different situations and make different emphases. The first interpretation might be the most natural one in terms of its human-centric description, which emphasizes that it is Bob who drives the car. From this point of view, we notice that Bob has to have the ability to compose prompts from the safety system and operate the car. However, it is also possible to claim it is the safety system that is using Bob as a tool to control the car.  This view emphasizes the system's ability to interact with humans and the car. The third interpretation considers both Bob and the safety system as inputs. This interpretation moves our attention to the car's ability to compose the driver and the safety system, which is related to its cockpit and interfaces to hold the safety system
\footnote{One reason the third interpretation might sound unnatural is that while Bob could drive other cars with other safety systems, and the safety system could operate with other drivers, there is less reason to think that the car could be driven safely by other systems and drivers. Thus, the three interpretations suggest different counterfactual judgments.}.

The above discussions show how choice of subject provides different insights into understanding. Our next question is how to leverage this flexibility in analyzing understanding in general. Though our framework is independent of the properties of the subjects, we will show how the structure of a subject can be revealed by choosing parts of the subjects as assistance (catalyst) of understanding. We will also demonstrate how the catalyst of understanding is related to concepts such as explanation and knowledge, which are regarded as important to understanding in previous theories. Finally, we will show how to use the notion of catalysts to clarify the acquisition of understanding, where a better understanding is formed.

\subsection{Catalyst}

Humans' understanding can be boosted with assistance. The previous example of understanding safe driving is not uncommon. 
Diagrams, such as that in Figure~\ref{fig:house}, can also be regarded as assistance as they help readers understand a novel more deeply.
More theoretically, consider the following example:
\begin{quote}
\textbf{(Proofs as catalysts)} The Traveling Salesman Problem (TSP) \cite{biggs1986graph} is an NP problem that asks whether there exists a path shorter than $n$ that traverses all nodes in an input graph. The problem is generally hard to solve, as the aforementioned TSP problem is an NP-complete problem for which people do not have an efficient algorithm. However, for a certain graph, if one can show an instance of a path shorter than $n$ (i.e. proof of the fact), answering the question can be done by verifying the instance (composing with the instance), which requires much fewer computational steps and could help the subject produce better outputs. 
\end{quote}
The path in the example can be regarded as a catalyst that boosts the subject's understanding in solving the TSP problem in the graph. \footnote{The idea of proofs as catalysts can be generalized to interactive provers \cite{goldwasser2019knowledge}, who talk with the subject and enhance the subject's understanding of the statement they want to prove. This makes catalysts not necessarily the answer to the problem. For example, a zero-knowledge prover of the TSP problem can prove the existence of a path without revealing it.
}

Based on the universal existence of assistance in all the above examples, we formally define a catalyst as follows:

\begin{definition}[Catalyst]
    Suppose a subject understands $O$ with the set $S$, in the view of verifier $V$ (with $S^*_V$). In the view of  $V$, $C$ is a catalyst of understanding if it satisfies that, for each element $\vec{I}=[I_1,I_2,...]$ of $S^*_V$, composing $\vec{I}'=[C,I_1,I_2,...]$ and use the output $I_{\rm out}^*$ to replace the original $I_{\rm out}$ in each $(\vec{I}, I_{\rm out})\in S$, makes a new set $S'$ that is more favourable in the view of $V$.
\end{definition}

People might find the definition of catalyst too loose as it can include nearly everything that helps to understand. For example, a neat workplace can be considered a catalyst because it helps the worker work better and produce more favourable $S'$. Though the workplace environment seems non-intellectual, we still regard it as a catalyst due to the minimal nature of our framework - it ignores the properties of the object involved besides its effect in changing understanding. We note that the purpose of defining catalysts is to help the user of our framework find important catalysts that matter in analyzing a situation of understanding. In the workplace example, gravity can also be regarded as a catalyst because gravity is necessary for the workers to carry out their work. However, noticing gravity as a catalyst, in this case, is less useful than noticing the effect of tidiness. Similarly, tidiness might be less important than intellectual catalysts such as well-written work instructions. The usefulness of a catalyst could depend on factors such as reusability, transferability, and associated cognitive costs such as storage size and demands of use.

The notion of catalyst can readily apply to the design of LLM-based AI systems. Designing catalysts for LLMs to use is an area of active research. To enhance the performance of LLMs, various methods for designing the prompts that are input to them have been proposed. For example, in chain-of-thought prompting \cite{wei2022chain}, LLMs are prompted to carry out multi-round reasoning, enabling them to address complex inquiries without necessitating alterations to the LLMs themselves. 

Retrieval augmented generation (RAG) \cite{lewis2020retrieval} is another example. To help LLMs use knowledge from a long document, a common approach is to break it into fragments and retrieve the relevant fragments based on the context. The retrieved fragments will be input to the LLM along with other inputs to produce better outputs. The fragments retrieved can be regarded as direct catalysts. From another view, one can also regard the program that retrieves the fragments as a catalyst,  producing outputs similar to those of composing the entire long document when it is composed.

From the perspective of philosophy, many of the conditions of understanding proposed in previous work can be considered as requirements of catalysts. For example, many theorists believe that explanation is central to understanding \cite{kvanvig2003value, strevens2013no, khalifa2013role, krenn2022scientific}. In our framework, explanations can be viewed as catalysts that boost understanding because we can expect composing inputs related to the phenomenon with the explanation of the phenomenon would produce better outputs.
Extending this idea, we treat explanation-seeking curiosity \cite{liquin2020explanation} as a kind of desire for catalysts.

On a larger scale, scientific theories can also be seen as catalysts when treating them as tools that generate explanations or solve problems. Kuhn considers scientific paradigms as instruments that help scientists solve scientific puzzles and regards outstanding puzzle-solving ability as the core reason new paradigms are accepted \cite{kuhn2012structure, sep-thomas-kuhn}. In our framework, this is equivalent to saying that scientific paradigms are catalysts for scientific understanding, and paradigms are accepted if they help compose important inputs that have not been properly composed by their predecessors.
In the context of scientific understanding, De Regt proposes that understanding a phenomenon requires fitting it into a broader theory \cite{de2005contextual}. We claim that the requirement of a theory can be seen as a requirement of a catalyst, which proves the subject has the ability to properly compose a broader scope of inputs besides the standalone phenomenon. 

Finally, we note that catalysts do not have to be directly related to the object of understanding though we defined so, and catalysts could be applied across domains. For example, in the case of analogical reasoning, having a theory of motion in space can improve understanding of the analogically linked case of ``motion'' of the solution moving in the parameter space when using gradient descent methods based on momentum and friction \cite{qian1999momentum}. More generally, any bits of information, such as a vivid across-domain metaphor, that makes the explanation more accessible will also count as catalysts. 

\subsection{Subject decomposition}

\begin{figure*}
    \centering
    \includegraphics[width=1\linewidth]{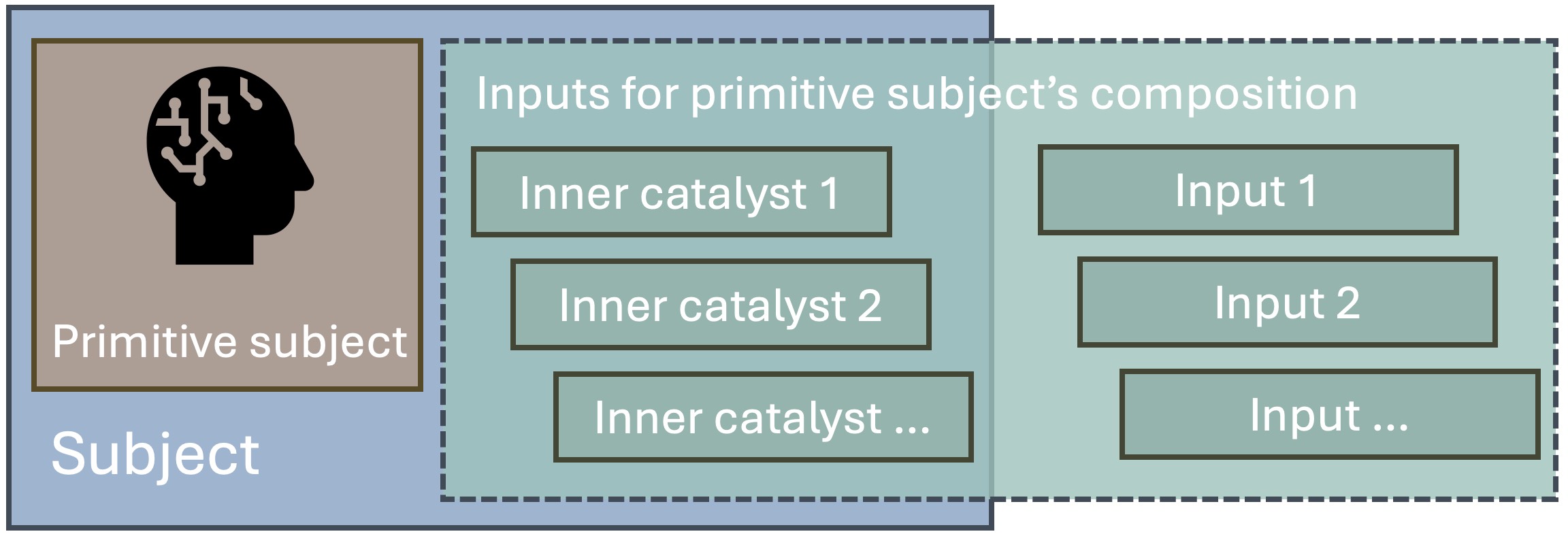}
    \caption{Subject decomposition for a general subject. We reveal a subject's structure by decomposing it into inner catalysts and a primitive subject so that the composition done by the subject can be viewed as the primitive subject composing inner catalysts and inputs from outside. The learning ability of the subject is determined by its ability to produce inner catalysts.}
    \label{fig:subject-decomp}
\end{figure*}

When we assess a subject's understanding by mere composability, we treat subjects as black boxes, not considering their internal structures. However, most subjects have an internal structure, which matters for assessing understanding. For example, in the example of Bob driving the car, the composite subject can be naturally decomposed into the car, Bob, and the safety system. Given a subject and a catalyst, we can always treat them as a whole, and this composite subject's internal structure would include the original subject and the catalyst. The examples above show how we make a composite subject and decompose it to identify its internal structure. 

A similar idea can also be applied when we want to decompose a subject whose structure is unclear. For example, when we say Charlie understands the sky is blue. It is common to say that Charlie should be able to use \textbf{his knowledge} of the sky's colour. Theoretically, Charlie's knowledge is a part of Charlie's body, which all functions in a black-box way to produce the output. However, in some contexts, we find it helpful to decompose Charlie into a subject that can process knowledge and then some knowledge about the sky, optics, etc. This decomposition allows us to reason about the role of each decomposed part in forming the understanding by acting as a ``catalyst'' for understanding. 
It helps us to ask and answer ``if he did not have this part, what would be his understanding?'', ``what caused him to get this part?'', ``how well can he use this part?'', etc. With the above heuristics, we define the \textbf{inner catalyst} below as a central concept to decompose and find the structure of a subject.

\begin{definition}[Subject decomposition]
Regarding a subject's understanding of the object $O$, $C$ is an \textbf{inner catalyst} if 1. $C$ is a part of the subject; 2. $C$ can be regarded as a catalyst that is composed by a \textbf{primitive subject} who is a part of the subject. 
\end{definition}
The following example shows an application of subject decomposition.
\begin{quote}
\textbf{(Optimizing neural networks)} To see how the understanding of a neural network can be improved, we can first decompose it into two inner catalysts held by the backbone computer: the network architecture and the corresponding network parameters. Based on the decomposition, it is natural to see that we can improve its understanding by 1.~optimizing the parameters, 2.~finding better network architecture, and 3.~fabricating better computers that support a better network architecture.
\end{quote}

Subject decomposition encourages us to find a proper set of inner catalysts and study their interaction with the primitive subject. At the beginning of Section~\ref{sec:structure}, we discussed how we can describe the understanding of safe driving in the composite subject in different ways. With subject decomposition, we see that Bob and the safety system can be seen as two straightforward examples of the inner catalysts when treating the car as a primitive subject. This decomposition, thus, reminds us to think about the different roles that Bob and the safety system take when composing inputs, as well as the primitive subject's (the car's) ability to use them as catalysts. Subject decomposition can also be a tool for extracting cognitive architecture in biological subjects. As shown in the example of Charlie, explanations, knowledge, and theories that are stored in a person's mind are also useful inner catalysts to discuss when analyzing the subject's understanding. Though these catalysts might be totally imagined by us as they are usually stored in an undifferentiated way, treating them as inner catalysts still helps to clarify the basis of the instance of understanding and make predictions about the improvement and decay of understanding in other circumstances.

As an application of subject decomposition, ablation study is a widely adopted strategy for investigating the structure of AI systems \cite{meyes2019ablation}. This technique involves systematically disabling or removing individual components of the AI system, such as a structure in the neural network, to evaluate their contribution to the overall performance. We note that it is necessary to find a proper decomposition of the system to disable its components in this process.
From the view of subject decomposition, we regard this process as decomposing the system into inner catalysts and testing how each catalyst contributes to the AI system's understanding by observing the change of understanding when altering the catalysts.

\subsection{Acquisition of understanding}

An important reason we propose subject decomposition is that it allows us to reason about the elements constituting the subject concerning understanding and, therefore, helps analyze the acquisition of the subject's understanding. For example, it makes it possible to ask questions such as ``how do changes in inner catalysts cause the improvement of understanding?'' and conversely, ``how is the improvement of understanding related to the changes of inner catalysts?''. If the car with Bob appears to drive more safely, the decomposition helps us conjecture that it might be either the safety system, Bob, or the car itself is improved. It also lets us consider how improving Charlie's knowledge can enhance his response when answering questions posed by others. 

However, is it always possible to reason about the acquisition of understanding by the notion of catalysts? Our answer is yes. We note that there are only two cases when the understanding of a subject is enhanced when the understood object $O$ is unchanged. The first is when a new catalyst is added to the composition, and the second is when the subject is improved. The first case is trivially related to catalysts. For the second, note that whenever we attribute the acquisition of understanding to the improvement of the subject, we can always attribute to the improvement of a part of the subject.
As the improved part improves understanding,  we can treat it as a catalyst that is inside the subject. We can also use an example to demonstrate the relationship between the acquisition of understanding and catalysts.
\begin{quote}
\textbf{(Improving understanding)} To better understand physics, Alice can use a better catalyst, such as a textbook or search engine, when she composes physics-related inputs. Besides, she must update herself by updating her inner catalysts, such as any knowledge that is related to physics, as well as her physiological conditions, such as blood sugar level, that affect her overall thinking ability.
\end{quote}
We summarize the above arguments as the following claim.
\begin{claim}[Acquisition of understanding]
According to verifier $V$, the subject must obtain and employ a new catalyst (inner or external) to make a subject's understanding more favourable to $V$.
\end{claim}

The above claims show how the acquisition of understanding is related to catalysts. However, it is worth pointing out that the acquisition of understanding is not equivalent to learning, as learning seems to entail the subject improving her understanding at least in part by herself -- learning is a success attributable to a subject, as opposed to mere change which can happen accidentally or by outside pressure. 
For example, when the search engine provides better results, though Alice might show better understanding by using it, it sounds unnatural that Alice demonstrated her learning ability through the search engine updates. Nonetheless, by updating her knowledge by reading textbooks, Alice demonstrates her ability to make new inner catalysts, and it is intuitive that Alice learned in this process. With the notion of catalysts, we can define the learning ability of a subject as below.

\begin{claim}[Learning ability]
\label{learning-ability}
A subject's learning ability is determined by its ability to compose related inputs into an update of the subject's inner catalysts.
\end{claim}

We note that the learning ability of a subject is of special importance. If a subject's learning ability is absent, it must wait for an external process to improve. For example, imagine there is a computer vision model that gives categories to pictures that wrongly recognize a panda as a gibbon. The model can be improved by some external forces, such as fine-tuning by humans. However, the model is far less intelligent than humans as their understanding acquisition depends heavily on other subjects. In contrast, humans can accept inputs (such as ``this is a panda, not a gibbon'', along with an image of a panda) and update their inner catalysts by themselves, allowing them to learn to achieve higher understanding gradually.

\section{General intelligence and learning}

We have developed theories to characterize a subject's understanding and analyze the structure behind the understanding by catalysts.
An immediate question that follows is: ``how could we apply this framework to the current LLM-based systems and measure their distance towards the `general intelligence' of humans?''
It might be tempting to regard problems, such as hallucinations, failures on counterfactual tasks, and prompt brittleness, as the gap between LLM-based systems and humans. However, it is obvious that humans also suffer from similar problems. It is also common for humans to be overconfident about their knowledge, unable to generalize their ability to new tasks, and be too sensitive to inputs. 

Instead of noticing the obvious gaps between the current AI systems and humans, we compare them at a higher level by their learning ability.
Humans have strong learning abilities, which might make learning ability an indispensable part of general intelligence in a human verifier's view. 
More importantly, no matter how the AI system understands at a point, a high-level learning ability enables the system to improve its understanding to the level that satisfies the verifier. When evaluating whether the AI system has achieved general intelligence, humans usually only expect that the system has acquired a level of understanding that is similar to humans, and humans usually know how this level of understanding can be acquired by human-style learning, such as reading a specific textbook and remembering some key facts. In all the problems we mentioned for LLMs, though humans also suffer from similar problems, we usually do not think they are severe drawbacks. We claim that a crucial reason is that we know how humans can overcome these problems by learning. This point is revealed by the following analysis.

\begin{quote}
\textbf{(Baby intelligence)}
We usually regard human babies as having general intelligence, though their understanding of many objects is low when born. This is because we know that they have human-level learning abilities and will acquire human-level understanding with their learning abilities 
(either innate or acquired over development).
\end{quote}

Though current AI systems have demonstrated learning ability to some extent, it is still unclear how human-level artificial learners can be built, and it is beneficial to investigate the major gaps between the learning ability of current AI systems and humans.

\subsection{Gaps in learning ability}

We have formulated learning ability as the ability to compose inputs into the subject's inner catalysts in Claim~\ref{learning-ability}, which allows us to analyze learning ability by composability. In fact, we can also view learning ability as an instance of understanding and use the same framework to characterize it. 
This makes salient the universality and scale of learning ability.

Humans have the ability to handle universal types of inputs for learning. For example, students would build their knowledge during classes through texts, diagrams, and the teacher's voices. In contrast, neural network-based learning only accepts a narrow format of inputs in general. For example, neural networks for image classification usually only take pairs of images and categories as inputs. It is impossible for them to learn from natural language instructions such as ``Pandas are bear-like animals with black and white hair''. A similar problem also happens in the training of LLMs. Though LLMs understand natural language instructions, there is no established way to optimize their parameters following these instructions.

The scale of learning ability also takes a critical role in its characterization. Humans can learn from long textbooks, movies, and audiobooks. It seems neural networks can learn from the larger scale of inputs as their training sets can easily exceed the scale humans can handle. However, it is important to note that neural networks do not treat these inputs similarly to humans due to their lack of universality in processing the inputs. When optimizing the parameters of LLMs by corpus, the texts are only interpreted as meaningless tokens that are sampled from a probability distribution. Instead, when humans read textbooks, they can take them as inputs with semantic meanings and follow the instructions in them to build their knowledge. Especially, humans can relate new knowledge to old knowledge and produce a coherent knowledge network, which boosts their performance in large-scale learning. In contrast, it is currently still unclear how to accurately update the knowledge inside LLMs \cite{yao2023editing, hoelscher2023detecting, zhang2024comprehensive}, let alone enforce the coherence of their knowledge.

\subsection{Autocatalysis and learning}

All these gaps between human and machine learners indicate there is still huge space for AI to improve further. Though LLMs are not a complete solution to it, we still argue that the advent of LLMs brings us closer to the goal of general intelligence. Our argument is based on the following special property of LLMs that is rare in other machine-learning models\footnote{We note that machine-learning models are not equal to machine-learning systems, which include the models' training programs. In this context, we regard machine-learning models as the products of machine-learning systems. }.

\begin{observation}
LLMs can use their outputs as catalysts.
\end{observation}

We highlight that this \textbf{autocatalysis} property of LLMs cannot be found in the machine-learning models whose inputs and outputs are of different types, such as models for image recognition and image generation from texts. For models whose inputs and outputs are of the same type, such as natural language processing models for text summarization and translation \cite{khurana2023natural}, it is also not usual that their outputs can serve as catalysts to boost their performance.

However, it is very common that the outputs of a run of LLM are treated as part of the inputs of another run, and this multi-stage processing of inputs has been leveraged in many studies \cite{yao2022react, zhou2023language, yoshikawa2023large}. The outputs that are input to LLMs again can be treated as catalysts, as they usually carry critical information that helps produce better outputs and, therefore, demonstrate better understanding.

It is not hard to notice that the autocatalytic property of LLMs is closely related to the learning ability described. 
The autocatalytic property implies we can use LLMs as primitive subjects that can produce the subject's inner catalysts, allowing us to use LLMs' understanding as the foundation of the subject's learning ability for boosting its universality and scale.
For example, it is not hard to imagine LLMs using tools to accept inputs from types other than texts and produce inner catalysts based on them. There is also no obstacle for LLMs to follow instructions to operate their inner catalysts.

Learning from large-scale inputs might also pose challenges to humans. It might involve multiple rounds of reading and note-taking to make a researcher fully understand a long paper. Regarding LLMs, it is not difficult to let LLM-based agents write notes and store them for future retrieval. When asked to build knowledge, it is possible for LLMs to semantically search the existing knowledge and update the subject's knowledge based on it, which opens up the way for building coherent knowledge networks for large-scale inputs for learning. 
Various methods for processing, storing, and retrieving knowledge built from long inputs have been proposed in LLM-based systems \cite{sarthi2024raptor, chen2023walking, packer2023memgpt, gao2023retrieval}.
Though it is still complex for these LLM-based systems to achieve the same performance as humans in learning, we see great potential for improvement in the future.

\section{Conclusion}

We have proposed a practical and minimal framework to characterize understanding by the subject's behaviour in composing inputs into outputs. We demonstrated how to use this idea to analyze the universality and scale of the subject's understanding. Based on the framework, we established how to use the notion of catalysts to formulate assistance for understanding and how to reveal the structure of a subject by the notion of primitive subject and inner catalysts. Further, we argue that the learning ability of a subject is determined by its ability to compose inputs into its inner catalysts. We pointed out that a higher level of learning ability is critical for AI systems to reach general intelligence. Finally, we show how LLMs pave the way for implementing human-level artificial learners by their ability of autocatalysis.

From the view of this work, our frameworks in this paper can be regarded as catalysts that help understand AIs' understanding. Moreover, the readers demonstrate the universality and scale of their learning ability by reading and building knowledge about our theory.

For future work, we note that we only demonstrated a small number of examples to which our framework can be applied. Besides universality and scale, more important characteristics might be formulated using the notion of composability.
For example, studying creativity as a characteristic of understanding might be interesting.
Regarding catalysts and subject decomposition, we mainly focused on how to analyze the structure of existing subjects. 
However, describing how to compose existing subjects into a larger subject, such as human or AI community, is also important.
Moreover, we only discussed LLMs as an example of autocatalytic models. Nonetheless, studying other useful autocatalytic subjects, such as humans and Turing machines, and seeing how their understanding is related to their learning ability may be helpful.
Finally, implementing practical software systems that employ the concepts laid out in our paper is hopefully a step towards artificial general intelligence. We hope that these concepts also help us build better AI-driven scientists and engineers in the future.

\ifcsname arxiv\endcsname
\section*{Acknowledgments}
We thank Reza Hadisi, Cameron Yetman, Jixuan Yi, Jocelyn Wang, and Yannis Kevrekidis for their helpful comments. A.A.-G. thanks Anders G. Frøseth for his generous support. A.A.-G. also acknowledges the generous support of Natural Resources Canada and the Canada 150 Research Chairs program.
\fi
\bibliography{philosophy, ai}

\begin{thebibliography}{70}
\providecommand{\natexlab}[1]{#1}

\bibitem[{Anthropic(2024)}]{anthropic2024claude}
Anthropic. 2024.
\newblock The Claude 3 Model Family: Opus, Sonnet, Haiku.

\bibitem[{Baumberger, Beisbart, and Brun(2016)}]{baumberger2016understanding}
Baumberger, C.; Beisbart, C.; and Brun, G. 2016.
\newblock What is understanding? An overview of recent debates in epistemology and philosophy of science.
\newblock \emph{Explaining understanding}, 1--34.

\bibitem[{Bender and Koller(2020)}]{bender2020climbing}
Bender, E.~M.; and Koller, A. 2020.
\newblock Climbing towards NLU: On meaning, form, and understanding in the age of data.
\newblock In \emph{Proceedings of the 58th annual meeting of the association for computational linguistics}, 5185--5198.

\bibitem[{Biggs, Lloyd, and Wilson(1986)}]{biggs1986graph}
Biggs, N.; Lloyd, E.~K.; and Wilson, R.~J. 1986.
\newblock \emph{Graph Theory, 1736-1936}.
\newblock Oxford University Press.

\bibitem[{Bird(2022)}]{sep-thomas-kuhn}
Bird, A. 2022.
\newblock {Thomas Kuhn}.
\newblock In Zalta, E.~N., ed., \emph{The {Stanford} Encyclopedia of Philosophy}. Metaphysics Research Lab, Stanford University, {S}pring 2022 edition.

\bibitem[{Brown et~al.(2020)Brown, Mann, Ryder, Subbiah, Kaplan, Dhariwal, Neelakantan, Shyam, Sastry, Askell et~al.}]{brown2020language}
Brown, T.; Mann, B.; Ryder, N.; Subbiah, M.; Kaplan, J.~D.; Dhariwal, P.; Neelakantan, A.; Shyam, P.; Sastry, G.; Askell, A.; et~al. 2020.
\newblock Language models are few-shot learners.
\newblock \emph{Advances in neural information processing systems}, 33: 1877--1901.

\bibitem[{Browning and Lecun(2022)}]{browning_ai_2022}
Browning, J.; and Lecun, Y. 2022.
\newblock {AI} {And} {The} {Limits} {Of} {Language}.

\bibitem[{Chen et~al.(2023)Chen, Pasunuru, Weston, and Celikyilmaz}]{chen2023walking}
Chen, H.; Pasunuru, R.; Weston, J.; and Celikyilmaz, A. 2023.
\newblock Walking down the memory maze: Beyond context limit through interactive reading.
\newblock \emph{arXiv preprint arXiv:2310.05029}.

\bibitem[{Chowdhery et~al.(2023)Chowdhery, Narang, Devlin, Bosma, Mishra, Roberts, Barham, Chung, Sutton, Gehrmann et~al.}]{chowdhery2023palm}
Chowdhery, A.; Narang, S.; Devlin, J.; Bosma, M.; Mishra, G.; Roberts, A.; Barham, P.; Chung, H.~W.; Sutton, C.; Gehrmann, S.; et~al. 2023.
\newblock Palm: Scaling language modeling with pathways.
\newblock \emph{Journal of Machine Learning Research}, 24(240): 1--113.

\bibitem[{De~Regt and Dieks(2005)}]{de2005contextual}
De~Regt, H.~W.; and Dieks, D. 2005.
\newblock A contextual approach to scientific understanding.
\newblock \emph{Synthese}, 144: 137--170.

\bibitem[{Firt(2023)}]{firt2023artificial}
Firt, E. 2023.
\newblock Artificial understanding: A step toward Robust AI.
\newblock \emph{AI \& SOCIETY}, 1--13.

\bibitem[{Gao et~al.(2023)Gao, Xiong, Gao, Jia, Pan, Bi, Dai, Sun, and Wang}]{gao2023retrieval}
Gao, Y.; Xiong, Y.; Gao, X.; Jia, K.; Pan, J.; Bi, Y.; Dai, Y.; Sun, J.; and Wang, H. 2023.
\newblock Retrieval-Augmented Generation for Large Language Models: A Survey.
\newblock \emph{arXiv preprint arXiv:2312.10997}.

\bibitem[{Ge et~al.(2024)Ge, Hua, Mei, Tan, Xu, Li, Zhang et~al.}]{ge2024openagi}
Ge, Y.; Hua, W.; Mei, K.; Tan, J.; Xu, S.; Li, Z.; Zhang, Y.; et~al. 2024.
\newblock Openagi: When llm meets domain experts.
\newblock \emph{Advances in Neural Information Processing Systems}, 36.

\bibitem[{Goldwasser, Micali, and Rackoff(2019)}]{goldwasser2019knowledge}
Goldwasser, S.; Micali, S.; and Rackoff, C. 2019.
\newblock The knowledge complexity of interactive proof-systems.
\newblock In \emph{Providing sound foundations for cryptography: On the work of shafi goldwasser and silvio micali}, 203--225.

\bibitem[{Hannon(2021)}]{hannon2021recent}
Hannon, M. 2021.
\newblock Recent work in the epistemology of understanding.
\newblock \emph{American Philosophical Quarterly}, 58(3): 269--290.

\bibitem[{He et~al.(2015)He, Zhang, Ren, and Sun}]{he2015delving}
He, K.; Zhang, X.; Ren, S.; and Sun, J. 2015.
\newblock Delving deep into rectifiers: Surpassing human-level performance on imagenet classification.
\newblock In \emph{Proceedings of the IEEE international conference on computer vision}, 1026--1034.

\bibitem[{Hoelscher-Obermaier et~al.(2023)Hoelscher-Obermaier, Persson, Kran, Konstas, and Barez}]{hoelscher2023detecting}
Hoelscher-Obermaier, J.; Persson, J.; Kran, E.; Konstas, I.; and Barez, F. 2023.
\newblock Detecting Edit Failures In Large Language Models: An Improved Specificity Benchmark.
\newblock \emph{arXiv preprint arXiv:2305.17553}.

\bibitem[{Ji et~al.(2023{\natexlab{a}})Ji, Qiu, Chen, Zhang, Lou, Wang, Duan, He, Zhou, Zhang et~al.}]{ji2023ai}
Ji, J.; Qiu, T.; Chen, B.; Zhang, B.; Lou, H.; Wang, K.; Duan, Y.; He, Z.; Zhou, J.; Zhang, Z.; et~al. 2023{\natexlab{a}}.
\newblock Ai alignment: A comprehensive survey.
\newblock \emph{arXiv preprint arXiv:2310.19852}.

\bibitem[{Ji et~al.(2023{\natexlab{b}})Ji, Lee, Frieske, Yu, Su, Xu, Ishii, Bang, Madotto, and Fung}]{ji2023survey}
Ji, Z.; Lee, N.; Frieske, R.; Yu, T.; Su, D.; Xu, Y.; Ishii, E.; Bang, Y.~J.; Madotto, A.; and Fung, P. 2023{\natexlab{b}}.
\newblock Survey of hallucination in natural language generation.
\newblock \emph{ACM Computing Surveys}, 55(12): 1--38.

\bibitem[{Khalifa(2012)}]{Khalifa2012-KHAIUO}
Khalifa, K. 2012.
\newblock Inaugurating Understanding or Repackaging Explanation?
\newblock \emph{Philosophy of Science}, 79(1): 15--37.

\bibitem[{Khalifa(2013)}]{khalifa2013role}
Khalifa, K. 2013.
\newblock The role of explanation in understanding.
\newblock \emph{The British Journal for the Philosophy of Science}.

\bibitem[{Khurana et~al.(2023)Khurana, Koli, Khatter, and Singh}]{khurana2023natural}
Khurana, D.; Koli, A.; Khatter, K.; and Singh, S. 2023.
\newblock Natural language processing: State of the art, current trends and challenges.
\newblock \emph{Multimedia tools and applications}, 82(3): 3713--3744.

\bibitem[{Krenn et~al.(2022)Krenn, Pollice, Guo, Aldeghi, Cervera-Lierta, Friederich, dos Passos~Gomes, H{\"a}se, Jinich, Nigam et~al.}]{krenn2022scientific}
Krenn, M.; Pollice, R.; Guo, S.~Y.; Aldeghi, M.; Cervera-Lierta, A.; Friederich, P.; dos Passos~Gomes, G.; H{\"a}se, F.; Jinich, A.; Nigam, A.; et~al. 2022.
\newblock On scientific understanding with artificial intelligence.
\newblock \emph{Nature Reviews Physics}, 4(12): 761--769.

\bibitem[{Kuhn(2012)}]{kuhn2012structure}
Kuhn, T.~S. 2012.
\newblock \emph{The structure of scientific revolutions}.
\newblock University of Chicago press.

\bibitem[{Kvanvig(2003)}]{kvanvig2003value}
Kvanvig, J.~L. 2003.
\newblock \emph{The value of knowledge and the pursuit of understanding}.
\newblock Cambridge university press.

\bibitem[{Lenci(2023)}]{lenci2023understanding}
Lenci, A. 2023.
\newblock Understanding natural language understanding systems. A critical analysis.
\newblock \emph{arXiv preprint arXiv:2303.04229}.

\bibitem[{Lewis et~al.(2020)Lewis, Perez, Piktus, Petroni, Karpukhin, Goyal, K{\"u}ttler, Lewis, Yih, Rockt{\"a}schel et~al.}]{lewis2020retrieval}
Lewis, P.; Perez, E.; Piktus, A.; Petroni, F.; Karpukhin, V.; Goyal, N.; K{\"u}ttler, H.; Lewis, M.; Yih, W.-t.; Rockt{\"a}schel, T.; et~al. 2020.
\newblock Retrieval-augmented generation for knowledge-intensive nlp tasks.
\newblock \emph{Advances in Neural Information Processing Systems}, 33: 9459--9474.

\bibitem[{Lewis(2016)}]{Tanya2016AlphaGo}
Lewis, T. 2016.
\newblock An AI expert says Google's Go-playing program is missing 1 key feature of human intelligence.
\newblock Accessed on May 23, 2024.

\bibitem[{Li et~al.(2023{\natexlab{a}})Li, Shao, Xie, Sheng, Zheng, Gonzalez, Stoica, Ma, and Zhang}]{li2023long}
Li, D.; Shao, R.; Xie, A.; Sheng, Y.; Zheng, L.; Gonzalez, J.; Stoica, I.; Ma, X.; and Zhang, H. 2023{\natexlab{a}}.
\newblock How Long Can Context Length of Open-Source LLMs truly Promise?
\newblock In \emph{NeurIPS 2023 Workshop on Instruction Tuning and Instruction Following}.

\bibitem[{Li et~al.(2023{\natexlab{b}})Li, Hammoud, Itani, Khizbullin, and Ghanem}]{li2023camel}
Li, G.; Hammoud, H. A. A.~K.; Itani, H.; Khizbullin, D.; and Ghanem, B. 2023{\natexlab{b}}.
\newblock CAMEL: Communicative agents for" mind" exploration of large language model society.
\newblock In \emph{Thirty-seventh Conference on Neural Information Processing Systems}.

\bibitem[{Li et~al.(2024)Li, Zhang, Do, Yue, and Chen}]{li2024long}
Li, T.; Zhang, G.; Do, Q.~D.; Yue, X.; and Chen, W. 2024.
\newblock Long-context llms struggle with long in-context learning.
\newblock \emph{arXiv preprint arXiv:2404.02060}.

\bibitem[{Liquin and Lombrozo(2020)}]{liquin2020explanation}
Liquin, E.~G.; and Lombrozo, T. 2020.
\newblock Explanation-seeking curiosity in childhood.
\newblock \emph{Current Opinion in Behavioral Sciences}, 35: 14--20.

\bibitem[{Liu et~al.(2023)Liu, Lin, Hewitt, Paranjape, Bevilacqua, Petroni, and Liang}]{liu2023lost}
Liu, N.~F.; Lin, K.; Hewitt, J.; Paranjape, A.; Bevilacqua, M.; Petroni, F.; and Liang, P. 2023.
\newblock Lost in the middle: How language models use long contexts.
\newblock \emph{arXiv preprint arXiv:2307.03172}.

\bibitem[{Lu et~al.(2022)Lu, Bartolo, Moore, Riedel, and Stenetorp}]{lu-etal-2022-fantastically}
Lu, Y.; Bartolo, M.; Moore, A.; Riedel, S.; and Stenetorp, P. 2022.
\newblock Fantastically Ordered Prompts and Where to Find Them: Overcoming Few-Shot Prompt Order Sensitivity.
\newblock In Muresan, S.; Nakov, P.; and Villavicencio, A., eds., \emph{Proceedings of the 60th Annual Meeting of the Association for Computational Linguistics (Volume 1: Long Papers)}, 8086--8098. Dublin, Ireland: Association for Computational Linguistics.

\bibitem[{Mandelkern and Linzen(2023)}]{mandelkern2023language}
Mandelkern, M.; and Linzen, T. 2023.
\newblock Do Language Models Refer?
\newblock \emph{arXiv preprint arXiv:2308.05576}.

\bibitem[{Meyes et~al.(2019)Meyes, Lu, de~Puiseau, and Meisen}]{meyes2019ablation}
Meyes, R.; Lu, M.; de~Puiseau, C.~W.; and Meisen, T. 2019.
\newblock Ablation studies in artificial neural networks.
\newblock \emph{arXiv preprint arXiv:1901.08644}.

\bibitem[{Min et~al.(2022)Min, Lyu, Holtzman, Artetxe, Lewis, Hajishirzi, and Zettlemoyer}]{min2022rethinking}
Min, S.; Lyu, X.; Holtzman, A.; Artetxe, M.; Lewis, M.; Hajishirzi, H.; and Zettlemoyer, L. 2022.
\newblock Rethinking the role of demonstrations: What makes in-context learning work?
\newblock \emph{arXiv preprint arXiv:2202.12837}.

\bibitem[{Mitchell and Krakauer(2023)}]{mitchell2023debate}
Mitchell, M.; and Krakauer, D.~C. 2023.
\newblock The debate over understanding in AI’s large language models.
\newblock \emph{Proceedings of the National Academy of Sciences}, 120(13): e2215907120.

\bibitem[{Nabokov(2017)}]{nabokov2017lectures}
Nabokov, V. 2017.
\newblock \emph{Lectures on Literature}.
\newblock Houghton Mifflin Harcourt.

\bibitem[{Nielsen(2016)}]{Michael2016AlphaGo}
Nielsen, M. 2016.
\newblock Is AlphaGo Really Such a Big Deal?
\newblock Accessed on May 23, 2024.

\bibitem[{OpenAI(2023{\natexlab{a}})}]{openai2023gpt4}
OpenAI. 2023{\natexlab{a}}.
\newblock GPT-4 Technical Report.
\newblock arXiv:2303.08774.

\bibitem[{OpenAI(2023{\natexlab{b}})}]{superalignment}
OpenAI. 2023{\natexlab{b}}.
\newblock Introducing Superalignment.
\newblock Accessed on May 23, 2024.

\bibitem[{Packer et~al.(2023)Packer, Fang, Patil, Lin, Wooders, and Gonzalez}]{packer2023memgpt}
Packer, C.; Fang, V.; Patil, S.~G.; Lin, K.; Wooders, S.; and Gonzalez, J.~E. 2023.
\newblock MemGPT: Towards LLMs as operating systems.
\newblock \emph{arXiv preprint arXiv:2310.08560}.

\bibitem[{Qian(1999)}]{qian1999momentum}
Qian, N. 1999.
\newblock On the momentum term in gradient descent learning algorithms.
\newblock \emph{Neural networks}, 12(1): 145--151.

\bibitem[{Qin et~al.(2023)Qin, Liang, Ye, Zhu, Yan, Lu, Lin, Cong, Tang, Qian et~al.}]{qin2023toolllm}
Qin, Y.; Liang, S.; Ye, Y.; Zhu, K.; Yan, L.; Lu, Y.; Lin, Y.; Cong, X.; Tang, X.; Qian, B.; et~al. 2023.
\newblock Toolllm: Facilitating large language models to master 16000+ real-world apis.
\newblock \emph{arXiv preprint arXiv:2307.16789}.

\bibitem[{Sarthi et~al.(2024)Sarthi, Abdullah, Tuli, Khanna, Goldie, and Manning}]{sarthi2024raptor}
Sarthi, P.; Abdullah, S.; Tuli, A.; Khanna, S.; Goldie, A.; and Manning, C.~D. 2024.
\newblock {RAPTOR}: Recursive Abstractive Processing for Tree-Organized Retrieval.
\newblock In \emph{The Twelfth International Conference on Learning Representations}.

\bibitem[{Shen et~al.(2023)Shen, Chen, Backes, and Zhang}]{shen2023chatgpt}
Shen, X.; Chen, Z.; Backes, M.; and Zhang, Y. 2023.
\newblock In chatgpt we trust? measuring and characterizing the reliability of chatgpt.
\newblock \emph{arXiv preprint arXiv:2304.08979}.

\bibitem[{Shi et~al.(2023)Shi, Chen, Misra, Scales, Dohan, Chi, Sch{\"a}rli, and Zhou}]{shi2023large}
Shi, F.; Chen, X.; Misra, K.; Scales, N.; Dohan, D.; Chi, E.~H.; Sch{\"a}rli, N.; and Zhou, D. 2023.
\newblock Large language models can be easily distracted by irrelevant context.
\newblock In \emph{International Conference on Machine Learning}, 31210--31227. PMLR.

\bibitem[{Silver et~al.(2016)Silver, Huang, Maddison, Guez, Sifre, Van Den~Driessche, Schrittwieser, Antonoglou, Panneershelvam, Lanctot et~al.}]{silver2016mastering}
Silver, D.; Huang, A.; Maddison, C.~J.; Guez, A.; Sifre, L.; Van Den~Driessche, G.; Schrittwieser, J.; Antonoglou, I.; Panneershelvam, V.; Lanctot, M.; et~al. 2016.
\newblock Mastering the game of Go with deep neural networks and tree search.
\newblock \emph{nature}, 529(7587): 484--489.

\bibitem[{Silver et~al.(2017)Silver, Schrittwieser, Simonyan, Antonoglou, Huang, Guez, Hubert, Baker, Lai, Bolton et~al.}]{silver2017mastering}
Silver, D.; Schrittwieser, J.; Simonyan, K.; Antonoglou, I.; Huang, A.; Guez, A.; Hubert, T.; Baker, L.; Lai, M.; Bolton, A.; et~al. 2017.
\newblock Mastering the game of go without human knowledge.
\newblock \emph{nature}, 550(7676): 354--359.

\bibitem[{Sosa(2010)}]{Sosa2010-SOSKFW-2}
Sosa, E. 2010.
\newblock \emph{Knowing Full Well}.
\newblock Princeton University Press.

\bibitem[{Strevens(2013)}]{strevens2013no}
Strevens, M. 2013.
\newblock No understanding without explanation.
\newblock \emph{Studies in history and philosophy of science Part A}, 44(3): 510--515.

\bibitem[{Sun et~al.(2021)Sun, Wang, Feng, Ding, Pang, Shang, Liu, Chen, Zhao, Lu et~al.}]{sun2021ernie}
Sun, Y.; Wang, S.; Feng, S.; Ding, S.; Pang, C.; Shang, J.; Liu, J.; Chen, X.; Zhao, Y.; Lu, Y.; et~al. 2021.
\newblock Ernie 3.0: Large-scale knowledge enhanced pre-training for language understanding and generation.
\newblock \emph{arXiv preprint arXiv:2107.02137}.

\bibitem[{Team et~al.(2023)Team, Anil, Borgeaud, Wu, Alayrac, Yu, Soricut, Schalkwyk, Dai, Hauth et~al.}]{team2023gemini}
Team, G.; Anil, R.; Borgeaud, S.; Wu, Y.; Alayrac, J.-B.; Yu, J.; Soricut, R.; Schalkwyk, J.; Dai, A.~M.; Hauth, A.; et~al. 2023.
\newblock Gemini: a family of highly capable multimodal models.
\newblock \emph{arXiv preprint arXiv:2312.11805}.

\bibitem[{Turing(2009)}]{turing2009computing}
Turing, A.~M. 2009.
\newblock \emph{Computing machinery and intelligence}.
\newblock Springer.

\bibitem[{Vaesen(2012)}]{vaesen2012cognitive}
Vaesen, K. 2012.
\newblock The cognitive bases of human tool use.
\newblock \emph{Behavioral and brain sciences}, 35(4): 203--218.

\bibitem[{Vinyals et~al.(2019)Vinyals, Babuschkin, Czarnecki, Mathieu, Dudzik, Chung, Choi, Powell, Ewalds, Georgiev et~al.}]{vinyals2019grandmaster}
Vinyals, O.; Babuschkin, I.; Czarnecki, W.~M.; Mathieu, M.; Dudzik, A.; Chung, J.; Choi, D.~H.; Powell, R.; Ewalds, T.; Georgiev, P.; et~al. 2019.
\newblock Grandmaster level in StarCraft II using multi-agent reinforcement learning.
\newblock \emph{Nature}, 575(7782): 350--354.

\bibitem[{Wang et~al.(2019)Wang, Pruksachatkun, Nangia, Singh, Michael, Hill, Levy, and Bowman}]{wang2019superglue}
Wang, A.; Pruksachatkun, Y.; Nangia, N.; Singh, A.; Michael, J.; Hill, F.; Levy, O.; and Bowman, S. 2019.
\newblock Superglue: A stickier benchmark for general-purpose language understanding systems.
\newblock \emph{Advances in neural information processing systems}, 32.

\bibitem[{Wang et~al.(2018)Wang, Singh, Michael, Hill, Levy, and Bowman}]{wang2018glue}
Wang, A.; Singh, A.; Michael, J.; Hill, F.; Levy, O.; and Bowman, S.~R. 2018.
\newblock GLUE: A multi-task benchmark and analysis platform for natural language understanding.
\newblock \emph{arXiv preprint arXiv:1804.07461}.

\bibitem[{Wang et~al.(2023)Wang, Ma, Feng, Zhang, Yang, Zhang, Chen, Tang, Chen, Lin et~al.}]{wang2023survey}
Wang, L.; Ma, C.; Feng, X.; Zhang, Z.; Yang, H.; Zhang, J.; Chen, Z.; Tang, J.; Chen, X.; Lin, Y.; et~al. 2023.
\newblock A survey on large language model based autonomous agents.
\newblock \emph{arXiv preprint arXiv:2308.11432}.

\bibitem[{Wei et~al.(2022)Wei, Wang, Schuurmans, Bosma, Xia, Chi, Le, Zhou et~al.}]{wei2022chain}
Wei, J.; Wang, X.; Schuurmans, D.; Bosma, M.; Xia, F.; Chi, E.; Le, Q.~V.; Zhou, D.; et~al. 2022.
\newblock Chain-of-thought prompting elicits reasoning in large language models.
\newblock \emph{Advances in Neural Information Processing Systems}, 35: 24824--24837.

\bibitem[{Wu et~al.(2023{\natexlab{a}})Wu, Bansal, Zhang, Wu, Zhang, Zhu, Li, Jiang, Zhang, and Wang}]{wu2023autogen}
Wu, Q.; Bansal, G.; Zhang, J.; Wu, Y.; Zhang, S.; Zhu, E.; Li, B.; Jiang, L.; Zhang, X.; and Wang, C. 2023{\natexlab{a}}.
\newblock Autogen: Enabling next-gen llm applications via multi-agent conversation framework.
\newblock \emph{arXiv preprint arXiv:2308.08155}.

\bibitem[{Wu et~al.(2023{\natexlab{b}})Wu, Qiu, Ross, Aky{\"u}rek, Chen, Wang, Kim, Andreas, and Kim}]{wu2023reasoning}
Wu, Z.; Qiu, L.; Ross, A.; Aky{\"u}rek, E.; Chen, B.; Wang, B.; Kim, N.; Andreas, J.; and Kim, Y. 2023{\natexlab{b}}.
\newblock Reasoning or reciting? exploring the capabilities and limitations of language models through counterfactual tasks.
\newblock \emph{arXiv preprint arXiv:2307.02477}.

\bibitem[{Xi et~al.(2023)Xi, Chen, Guo, He, Ding, Hong, Zhang, Wang, Jin, Zhou et~al.}]{xi2023rise}
Xi, Z.; Chen, W.; Guo, X.; He, W.; Ding, Y.; Hong, B.; Zhang, M.; Wang, J.; Jin, S.; Zhou, E.; et~al. 2023.
\newblock The rise and potential of large language model based agents: A survey.
\newblock \emph{arXiv preprint arXiv:2309.07864}.

\bibitem[{Yang et~al.(2023)Yang, Li, Wang, Lin, Azarnasab, Ahmed, Liu, Liu, Zeng, and Wang}]{yang2023mm}
Yang, Z.; Li, L.; Wang, J.; Lin, K.; Azarnasab, E.; Ahmed, F.; Liu, Z.; Liu, C.; Zeng, M.; and Wang, L. 2023.
\newblock Mm-react: Prompting chatgpt for multimodal reasoning and action.
\newblock \emph{arXiv preprint arXiv:2303.11381}.

\bibitem[{Yao et~al.(2022)Yao, Zhao, Yu, Du, Shafran, Narasimhan, and Cao}]{yao2022react}
Yao, S.; Zhao, J.; Yu, D.; Du, N.; Shafran, I.; Narasimhan, K.; and Cao, Y. 2022.
\newblock React: Synergizing reasoning and acting in language models.
\newblock \emph{arXiv preprint arXiv:2210.03629}.

\bibitem[{Yao et~al.(2023)Yao, Wang, Tian, Cheng, Li, Deng, Chen, and Zhang}]{yao2023editing}
Yao, Y.; Wang, P.; Tian, B.; Cheng, S.; Li, Z.; Deng, S.; Chen, H.; and Zhang, N. 2023.
\newblock Editing large language models: Problems, methods, and opportunities.
\newblock \emph{arXiv preprint arXiv:2305.13172}.

\bibitem[{Yoshikawa et~al.(2023)Yoshikawa, Skreta, Darvish, Arellano-Rubach, Ji, Bj{\o}rn~Kristensen, Li, Zhao, Xu, Kuramshin et~al.}]{yoshikawa2023large}
Yoshikawa, N.; Skreta, M.; Darvish, K.; Arellano-Rubach, S.; Ji, Z.; Bj{\o}rn~Kristensen, L.; Li, A.~Z.; Zhao, Y.; Xu, H.; Kuramshin, A.; et~al. 2023.
\newblock Large language models for chemistry robotics.
\newblock \emph{Autonomous Robots}, 47(8): 1057--1086.

\bibitem[{Zhang et~al.(2024)Zhang, Yao, Tian, Wang, Deng, Wang, Xi, Mao, Zhang, Ni et~al.}]{zhang2024comprehensive}
Zhang, N.; Yao, Y.; Tian, B.; Wang, P.; Deng, S.; Wang, M.; Xi, Z.; Mao, S.; Zhang, J.; Ni, Y.; et~al. 2024.
\newblock A comprehensive study of knowledge editing for large language models.
\newblock \emph{arXiv preprint arXiv:2401.01286}.

\bibitem[{Zhou et~al.(2023)Zhou, Yan, Shlapentokh-Rothman, Wang, and Wang}]{zhou2023language}
Zhou, A.; Yan, K.; Shlapentokh-Rothman, M.; Wang, H.; and Wang, Y.-X. 2023.
\newblock Language agent tree search unifies reasoning acting and planning in language models.
\newblock \emph{arXiv preprint arXiv:2310.04406}.

\end{thebibliography}

\appendix

\end{document}